\documentclass[10pt,twocolumn,letterpaper]{article}

\usepackage[applications]{wacv}
\usepackage{amsmath,amssymb}
\usepackage{booktabs}
\usepackage{multirow}
\usepackage{makecell}
\usepackage{placeins}
\usepackage{xspace}
\usepackage{graphicx}

\newcommand{\tablestyle}[2]{\setlength{\tabcolsep}{#1}\renewcommand{\arraystretch}{#2}\centering\small}

\newcommand{\method}{STAR\xspace}
\newcommand{\methodfull}{Selective Temporal Attenuation Regularization\xspace}

\definecolor{wacvblue}{rgb}{0.21,0.49,0.74}
\usepackage[pagebackref,breaklinks,colorlinks,allcolors=wacvblue]{hyperref}

\def\wacvPaperID{108}
\def\confName{WACV}
\def\confYear{2027}

\title{Tracking Is Not Sensing: Artifact-Aware Thermal Video Respiration Estimation in Head-Free Marmosets}

\author{
Penghao Wang$^{1,2,3}$ \quad
Sen Yang$^{1,2,3}$ \quad
Fenghua Xie$^{4}$ \quad
He Wang$^{1,3}$ \quad
Xindong Song$^{1,2,3,4}$\thanks{Corresponding author.}\\
$^{1}$Institute of Automation, Chinese Academy of Sciences\\
$^{2}$School of Artificial Intelligence, University of Chinese Academy of Sciences\\
$^{3}$State Key Laboratory of Brain Cognition and Brain-inspired Intelligence Technology\\
$^{4}$Tsinghua Laboratory of Brain and Intelligence, Tsinghua University\\
{\ttfamily\small \{wangpenghao2023, yangsen2023, he.wang, xindong.song\}@ia.ac.cn,}\\
{\ttfamily\small fenghuaxie@mail.tsinghua.edu.cn}
}

\begin{document}
\setcounter{footnote}{1} % The corresponding-author thanks uses the dagger symbol.
\maketitle
\begin{abstract}
Continuous physiological monitoring is essential for linking behavior, internal
state, and neural activity in neuroscience and translational research.
In common marmosets, a key nonhuman primate model for social neuroscience,
respiratory rate provides a basic readout of arousal, stress, and physiological
regulation.
Contact sensors can provide accurate measurements, but they add preparation
burden and may interfere with head-free behavioral protocols.
Thermal video offers a noncontact alternative by capturing breathing-related
temperature changes around the nose and mouth.
However, in head-free marmosets, these cues are weak, localized, and easily
corrupted by head motion, pose changes, and fur.
Here we show that this setting exposes a broader challenge in video-based
physiological sensing: visually reliable tracking does not necessarily imply
physiologically valid measurement.
Using synchronized thermal video and contact respiration recordings, we show
that marmoset respiration is recoverable from thermal video but strongly
ROI-sensitive.
Under head-free motion, visually plausible nasal tracking can still produce
traces inconsistent with contact respiration, demonstrating that tracking is not
sensing.
We therefore estimate respiratory rate from learned temporal signals extracted from thermal clips
and introduce \textbf{\methodfull(\method)}.
\method selectively attenuates high-motion thermal discontinuities through local
temporal imputation during training, regularizing models toward
respiration-consistent dynamics while leaving inference unchanged on raw video.
Across multiple deep-learning backbones, \method improves head-free estimation,
reaching 2.23 bpm MAE compared with 20.76 bpm for tracked ROI analysis.
\end{abstract}

\section{Introduction}
\label{sec:intro}

Continuous physiological monitoring is increasingly important for linking
behavior, internal state, and neural activity in naturalistic neuroscience and
translational research.
Physiological signals such as respiratory rate, heart rate, heart rate
variability, and surface temperature provide quantitative readouts of autonomic
regulation, arousal, stress, and behavioral state~\cite{appelhans2006heart,
shaffer2017overview,grossman2007toward,pavlidis2012fast,
molinaro2022contactless}.
In animal neuroscience, these measurements are especially valuable because
internal state can strongly modulate perception, action, and social behavior.
However, many physiological measurements still rely on contact sensors or
invasive instrumentation.
Although such devices provide accurate reference signals, their physical
attachment can add a preparation burden, disrupt behavior, and interfere with
head-free experimental protocols.
These limitations are particularly important for common marmosets, a small
nonhuman primate model widely used for social neuroscience and systems
neuroscience~\cite{miller2016marmosets}. In marmosets, respiratory and autonomic rhythms provide important readouts of physiological state and arousal-related behavior~\cite{kramer2019normal,bishop2022marmosetbreathing,borjon2016arousal}.

Thermal video offers a promising noncontact modality because breathing induces
periodic temperature changes around the nostrils, subnasal region, and
mouth~\cite{pereira2015breathing,pereira2017estimation,
mozafari2024thermalrr}.
Thermal imaging has also been used in marmosets to study nasal temperature
changes related to emotional state~\cite{ermatinger2019use}.
However, head-free marmoset respiration is a challenging vision problem with weak physiological signals:
the cue is small, spatially localized, and easily corrupted
by fur, pose changes, nostril orientation changes, and head motion.

This setting exposes a broader challenge in video-based physiological sensing:
visually reliable tracking does not necessarily imply physiologically valid
measurement.
Conventional thermal respiration pipelines localize a nasal or perinasal region
of interest (ROI), average its temperature over time, and estimate respiratory
rate from the dominant frequency~\cite{pereira2015breathing,
pereira2017estimation}.
These methods assume that a tracked ROI remains aligned with a stable
respiratory thermal source.
In head-free marmosets, this assumption can fail: a tracker may maintain a
visually plausible nasal region while the averaged thermal trace is dominated by
motion-induced thermal discontinuities rather than respiration.
In other words, tracking is not sensing.

We study this problem using synchronized thermal video and contact respiration
recordings from common marmosets under fixed-head and head-free conditions.
Fixed-head recordings show that respiration-related thermal information is
recoverable from video, but highly sensitive to ROI selection.
Head-free recordings reveal a stronger failure mode: even visually plausible
nasal tracking can produce traces inconsistent with contact respiration.
These observations motivate a shift from ROI-averaged tracking to learned spatiotemporal physiological signal extraction.

We therefore estimate respiratory rate from temporal signals predicted from thermal clips and adapt representative video-based physiological measurement backbones, originally developed for remote photoplethysmography, to recover respiratory waveforms from thermal sequences~\cite{chen2018deepphys,liu2020multi,
PhysNet,yu2022physformer}. To address motion-corrupted observations during training, we introduce
\textbf{{\methodfull} (\method)}, a motion-aware training strategy that uses
local temporal imputation to attenuate high-motion thermal discontinuities and
encourage respiration-consistent temporal dynamics.
Our contributions are summarized as follows:
\begin{itemize}
\item We study thermal respiratory monitoring in marmosets under
fixed-head and head-free conditions using synchronized thermal video
and contact respiration recordings.
\item We show that thermal respiration is highly ROI-sensitive:
under head motion, even visually plausible nasal tracking can fail
to preserve respiratory signals.
\item We introduce \method, a motion-aware training strategy that
improves head-free estimation across multiple backbones, achieving
2.23 bpm MAE versus 20.76 bpm for tracked ROI analysis.
\end{itemize}
\begin{figure*}[t]
    \centering
    \includegraphics[width=0.98\linewidth]{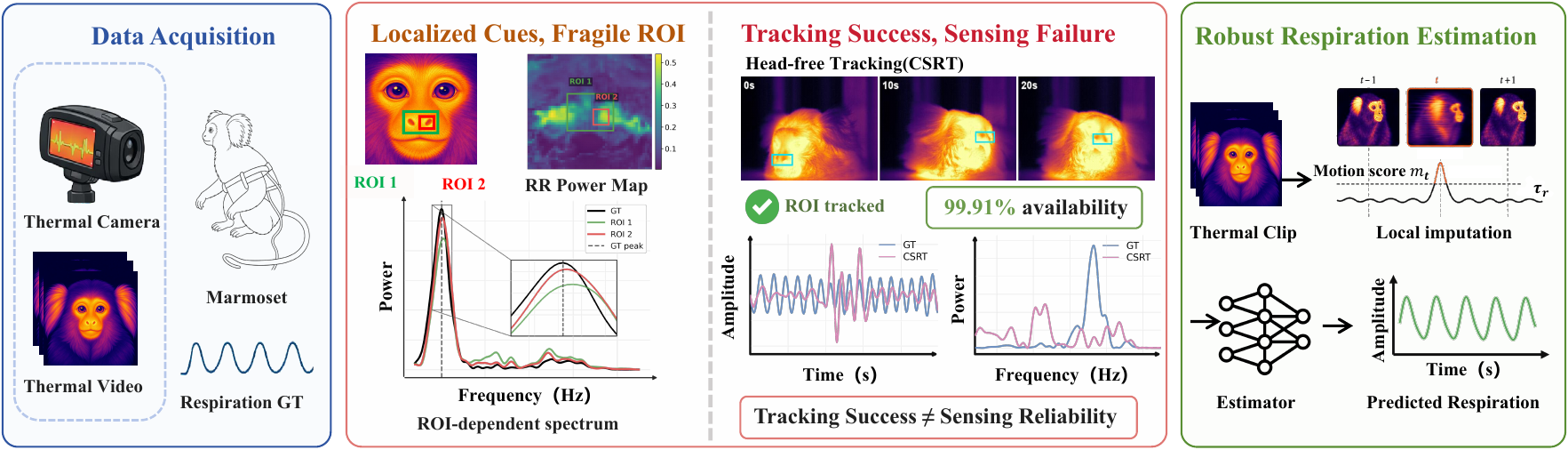}
    \caption{\textbf{Tracking is not sensing in head-free thermal respiration monitoring.}
    Left to right: synchronized thermal video and contact respiration provide
    paired observations; respiration related thermal cues are local and ROI
    sensitive; under head-free motion, visually successful CSRT tracking can still
    produce temperature traces that do not match respiration; robust estimation
    therefore requires recovering respiratory dynamics from thermal clips rather
    than relying on ROI averaged tracking alone.}
    \label{fig:overview}
\end{figure*}

\section{Related Work}
\label{sec:related}

\paragraph{Video based physiological measurement.}
Video based physiological measurement aims to estimate internal physiological
rhythms from image sequences without attached sensors.
Representative tasks include remote heart rate estimation, respiratory rate
estimation, and recovery of weak physiological waveforms from video.
Early studies mainly relied on handcrafted color projections, motion cues, or
signal decomposition to extract physiological traces from
video~\cite{verkruysse2008remote,poh2010advancements,
wang2016algorithmic}.
Recent methods have shifted toward learning based spatiotemporal models, where
neural networks recover physiological signals directly from video or estimate
vital rates from learned representations
~\cite{chen2018deepphys,liu2020multi,PhysNet,yu2022physformer,
liu2023efficientphys,narayanswamy2024bigsmall}.
These methods demonstrate that weak physiological information can be extracted
from visual signals.

\paragraph{Thermal video for respiratory monitoring.}
Thermal imaging is well suited for respiratory monitoring because respiration
induces periodic temperature variations around the nasal region.
Classical methods typically extract a temperature trace from a nasal
region of interest and analyze its temporal variations to estimate
respiratory rate~\cite{pereira2015breathing}.
Recent work has explored learning based approaches for thermal respiratory rate
estimation~\cite{mozafari2024thermalrr,pereira2017estimation}.
Beyond human studies, thermal imaging has also been applied to animal physiology
monitoring, including rodents, pigs, and wildlife
~\cite{barbosa2019contactless,pereira2018rodents,
sadeghi2024non,rzucidlo2023non}.
These studies demonstrate the feasibility of thermal respiration monitoring.
However, they are typically conducted in controlled settings or on larger
subjects.
This limits their applicability to marmoset behavior with large motion variation.
In such settings, small and highly localized respiratory regions make reliable
signal extraction particularly challenging.

\paragraph{Respiratory monitoring in head-free marmosets.}
Common marmosets are widely used in neuroscience and social behavior
research~\cite{miller2016marmosets}.
Their physiological rhythms are closely related to arousal, stress, and
behavioral state~\cite{borjon2016arousal,kramer2019normal}.
Recent vision tools have begun to support marmoset behavioral analysis, including
pose estimation~\cite{cheng2024marmopose}.
Thermal imaging has also been used to study marmoset emotional state through
nasal temperature variations~\cite{ermatinger2019use}.
However, continuous respiratory rate estimation from thermal video remains underexplored
in scenarios with large motion variation.
In particular, small respiratory regions, rapid motion, and pose changes make
ROI based thermal respiration estimation unreliable, motivating robust
respiratory signal recovery methods.

\section{From Localized Respiratory Cues to Tracking Failure}
\label{sec:data}

\subsection{Thermal Video Respiratory Rate Estimation}

We record synchronized thermal video and contact respiration from common
marmosets.
The thermal camera (MAG62 thermal imaging camera, Magnity Electronics Co., Ltd.)
captures $640 \times 512$ frames at 50 Hz and supports external triggering.
The contact respiration signal is sampled on the same hardware timed acquisition
system, so every thermal frame is aligned to the reference signal.
The full collection contains 48 five minute sessions from four male marmosets, for
240 minutes of synchronized thermal respiration data: 32 head-free sessions (160
minutes) and 16 fixed-head sessions (80 minutes).

For respiratory rate estimation, clip-level predictions are aggregated into a
20\,s analysis window following the frequency domain estimation protocol of
Cennini et al.~\cite{cennini2010heart}.
% Windows with severely corrupted respiratory signals due to motion artifacts are
% excluded by manual inspection to ensure reliable physiological measurement.
To ensure reliable physiological labels, windows are excluded by manual inspection
only when the reference respiration signal is severely corrupted,
such as by contact-sensor detachment or signal truncation.
This exclusion was based only on the quality of the reference respiration signal and was applied before evaluation.
The final evaluation set consists of 63 fixed-head and 150 head-free analysis
windows.
Examples of excluded and retained respiratory windows are provided in Appendix~\ref{sec:appendix_data_details}.

\subsection{Evaluation Protocol}

Let $\mathbf{X}=\{X_1,\ldots,X_T\}$ be a thermal clip, where $X_t \in
\mathbb{R}^{H\times W}$ is the $t$-th thermal frame.
For each analysis window, we estimate respiratory rate $\hat{r}$ in breaths per
minute and compare it with the contact reference $r$.
Traditional approaches compute an ROI averaged temperature trace from the input
clip, and estimate respiratory rate from its dominant frequency.
In contrast, learning based methods directly estimate a respiratory waveform
$\hat{s}(t)$ from the input clip, from which respiratory rate is derived using
the same frequency domain estimation protocol.

We evaluate respiratory rate error using mean absolute error (MAE), mean absolute
percentage error (MAPE), and Pearson correlation:
\begin{equation}
\mathrm{MAE}=\frac{1}{N}\sum_{i=1}^{N}|\hat{r}_i-r_i|,
\end{equation}
\begin{equation}
\mathrm{MAPE}=\frac{100}{N}\sum_{i=1}^{N}\left|\frac{\hat{r}_i-r_i}{r_i}\right|,
\end{equation}
\begin{equation}
\rho=
\frac{\sum_{i=1}^{N}(\hat{r}_i-\bar{\hat{r}})(r_i-\bar{r})}
{\sqrt{\sum_{i=1}^{N}(\hat{r}_i-\bar{\hat{r}})^2}
\sqrt{\sum_{i=1}^{N}(r_i-\bar{r})^2}}.
\end{equation}
% For key comparisons, we also report $R^2$ and Bland--Altman
% limits of agreement width.
% In head-free experiments, we additionally report success rates under 1, 3, and 5
% bpm error thresholds for motion stratified analysis.

Given a respiratory ROI $\Omega$, the frequency domain baseline computes the mean
thermal signal
\begin{equation}
s(t)=\frac{1}{|\Omega|}\sum_{(i,j)\in\Omega}X_t(i,j).
\end{equation}
After detrending, normalization, and band pass filtering, we estimate the power
spectral density $P(f)$ and choose the dominant frequency within the respiratory
band:
\begin{equation}
\hat{f}_{RR}=\arg\max_{f\in[f_{\min},f_{\max}]}P(f),\qquad \hat{r}=60\hat{f}_{RR}.
\end{equation}
This baseline is interpretable and computationally cheap, but it assumes that the
ROI stays aligned with a respiratory thermal source and that the ROI mean is
dominated by respiratory variation.

\subsection{Fixed Head Feasibility and ROI Sensitivity}

Fixed-head recordings provide a controlled setting to test whether marmoset
thermal video contains recoverable respiratory information.
ROI based frequency analysis indicates that respiration related thermal variation
can be measured when the region is appropriately localized, following classical
thermal respiration approaches that estimate breathing from temperature variation
in a nasal or perinasal ROI~\cite{pereira2015breathing}.
We first observe that respiratory rate estimation varies significantly across
different ROI selections around the nasal region, indicating strong spatial
sensitivity of the thermal signal.
To further analyze this sensitivity, we evaluate a post hoc oracle ROI that
serves as a nondeployable diagnostic upper bound for ROI frequency analysis.
The oracle ROI is obtained by minimizing respiratory rate error at the analysis
window level.
The search procedure is described in
Appendix~\ref{sec:appendix_protocol_details}.

\begin{figure}[t]
    \centering
    \makebox[\linewidth][c]{%
    \begin{subfigure}[t]{0.33\linewidth}
        \centering
        \includegraphics[width=\linewidth,height=0.80in,keepaspectratio]{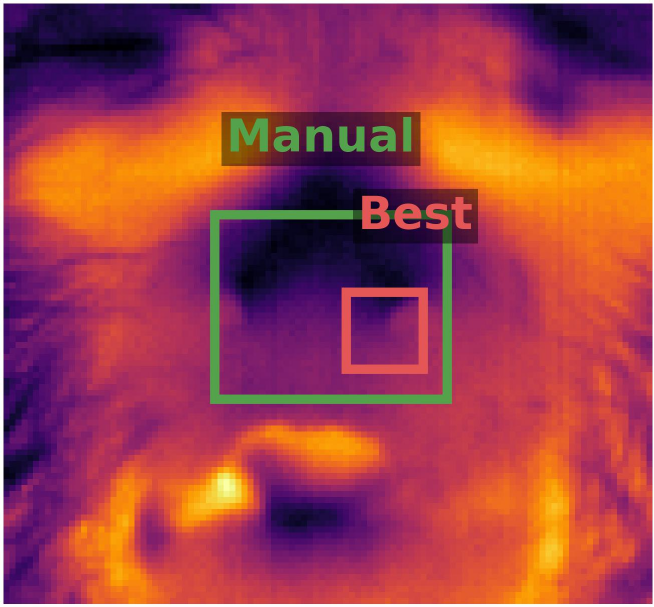}
        \caption{Thermal frame}
    \end{subfigure}\hspace{0.01\linewidth}%
    \begin{subfigure}[t]{0.33\linewidth}
        \centering
        \includegraphics[width=\linewidth,height=0.80in,keepaspectratio]{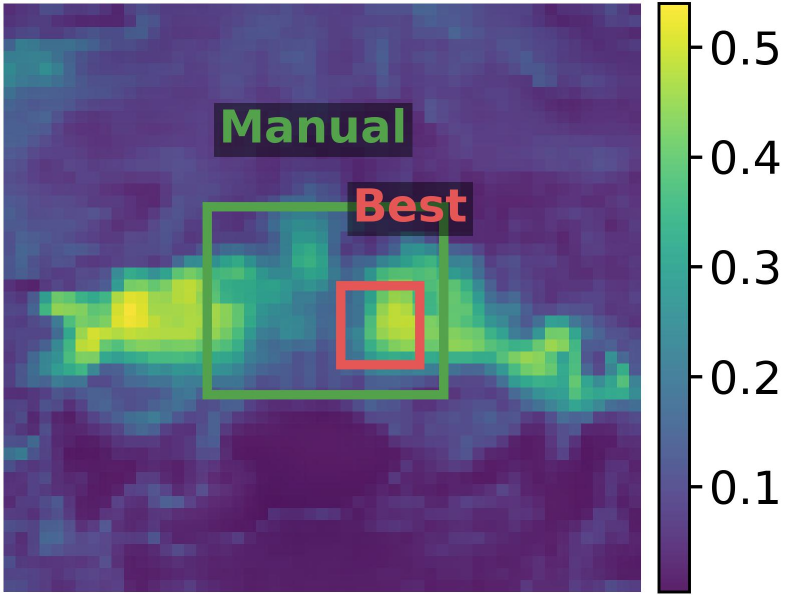}
        \caption{RR Power Map}
    \end{subfigure}\hspace{0.01\linewidth}%
    \begin{subfigure}[t]{0.33\linewidth}
        \centering
        \includegraphics[width=\linewidth,height=0.80in,keepaspectratio]{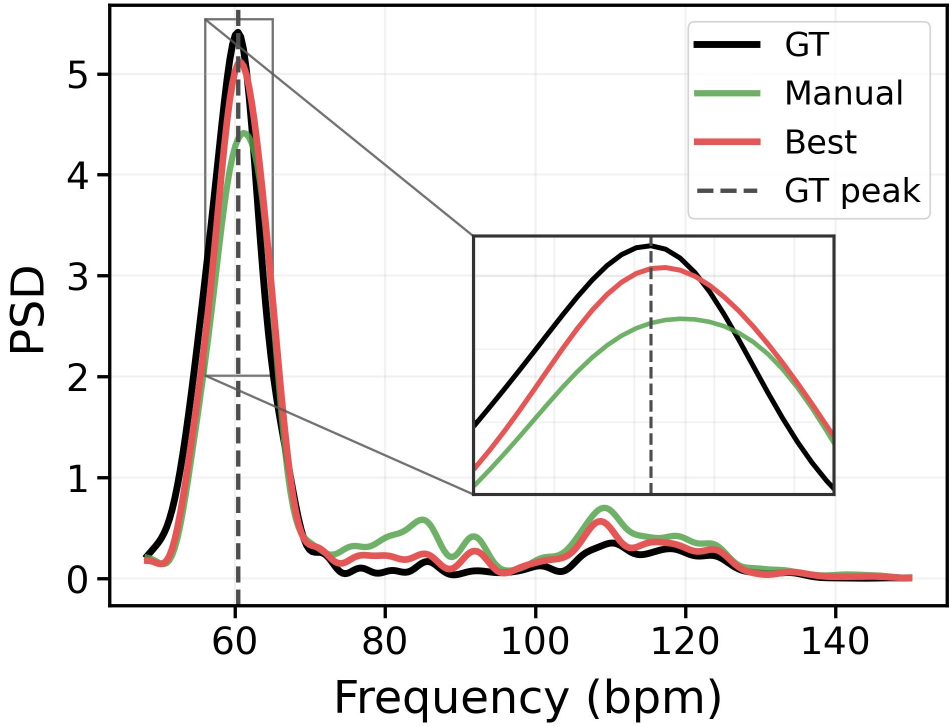}
        \caption{PSD}
    \end{subfigure}%
    }
    \caption{Manual ROI versus post-hoc oracle ROI under fixed-head conditions.}
    \label{fig:best_roi}
\end{figure}

As shown in Table~\ref{tab:fixed_roi}, the manual ROI achieves moderate accuracy,
while the oracle ROI significantly improves performance.
This gap indicates that the thermal respiratory signal is present but highly
sensitive to precise spatial localization.
Figure~\ref{fig:best_roi} further supports this finding.
The RR band power map shows that respiratory energy is strongly localized near
the nostril region.
A signal energy heatmap analysis (details in Appendix~\ref{sec:appendix_data_details})
further confirms this spatial sparsity,
showing that only a small region contributes most of the respiratory band power.
Compared with the manual ROI, the oracle ROI captures this concentrated region
more effectively, resulting in a sharper spectral peak in the power spectral
density (PSD) that is better aligned with the contact reference signal.
Overall, these results indicate that respiratory information is spatially
localized, and that estimation errors under fixed-head conditions are primarily
caused by imperfect spatial localization rather than the absence of a measurable
respiratory signal.

\begin{table}[t]
    \caption{Fixed-head ROI frequency analysis. The oracle ROI is a post hoc
        diagnostic upper bound and is not deployable.}
    \label{tab:fixed_roi}
    \tablestyle{8pt}{1.1}
    \begin{tabular}{lccc}
        \toprule
        ROI setting & MAE $\downarrow$ & MAPE $\downarrow$ & $\rho$ $\uparrow$ \\
        \midrule
        Manual ROI & 2.46 & 3.57\% & 0.87 \\
        Oracle ROI & \textbf{0.45} & \textbf{0.66\%} & \textbf{0.99} \\
        \bottomrule
    \end{tabular}
\end{table}

\subsection{Tracking Availability Does Not Ensure Respiratory Accuracy}

A natural extension of ROI based analysis is to track the respiratory region
over time under head-free motion.
We first evaluate Kernelized Correlation Filter (KCF)~\cite{henriques2015high},
following prior thermal respiratory monitoring work that used KCF for nose and eye
ROIs~\cite{sadeghi2024non}.
Since KCF often fails under marmoset head-free motion,
we further evaluate the Channel and Spatial Reliability Tracker
(CSRT)~\cite{lukezic2017discriminative} as a stronger tracker for comparison.
We evaluate tracked ROI baselines on 150 head-free analysis windows.
Each tracker is initialized from a manual perinasal ROI and updated frame by
frame.
CSRT achieves high visual continuity across most windows, where ROI availability
is defined in Appendix~\ref{sec:appendix_tracking_details}.
However, availability only measures tracking continuity and does not guarantee
that the ROI accurately covers the respiratory region.
We further perform manual verification to ensure consistent nasal region coverage
and retain 141 valid head-free analysis windows satisfying this criterion.
Even under this stricter criterion, respiratory rate estimation remains poor, as
shown in Table~\ref{tab:tracking}.
Additional tracking examples are provided in Appendix~\ref{sec:appendix_tracking_examples}.

\begin{table}[ht]
    \caption{Head-free tracking results. High ROI availability and visually
        correct nasal coverage do not translate into accurate respiratory-rate
        estimation.}
    \label{tab:tracking}
    \tablestyle{8pt}{1.1}
    \begin{tabular}{lcccc}
        \toprule
        Method & Windows & MAE $\downarrow$ & MAPE $\downarrow$ &
        $\rho$ $\uparrow$ \\
        \midrule
        KCF ROI & 150 & 21.68 & 21.80 & 0.06 \\
        CSRT ROI & 150 & 22.17 & 22.54 & 0.08 \\
        CSRT ROI & 141 & \textbf{20.76} & \textbf{21.55} & \textbf{0.10} \\
        \bottomrule
    \end{tabular}
\end{table}

These results expose the mismatch between visual tracking and physiological
sensing.
Although a tracker can maintain a plausible nasal bounding box, head motion
alters facial orientation, thermal distribution, and the proportion of respiratory
signal within the ROI.
As a result, the ROI mean becomes dominated by motion induced thermal variation
rather than respiration.

\section{Method}
\label{sec:method}
\begin{figure*}[ht]
    \centering
    \includegraphics[width=0.98\linewidth]{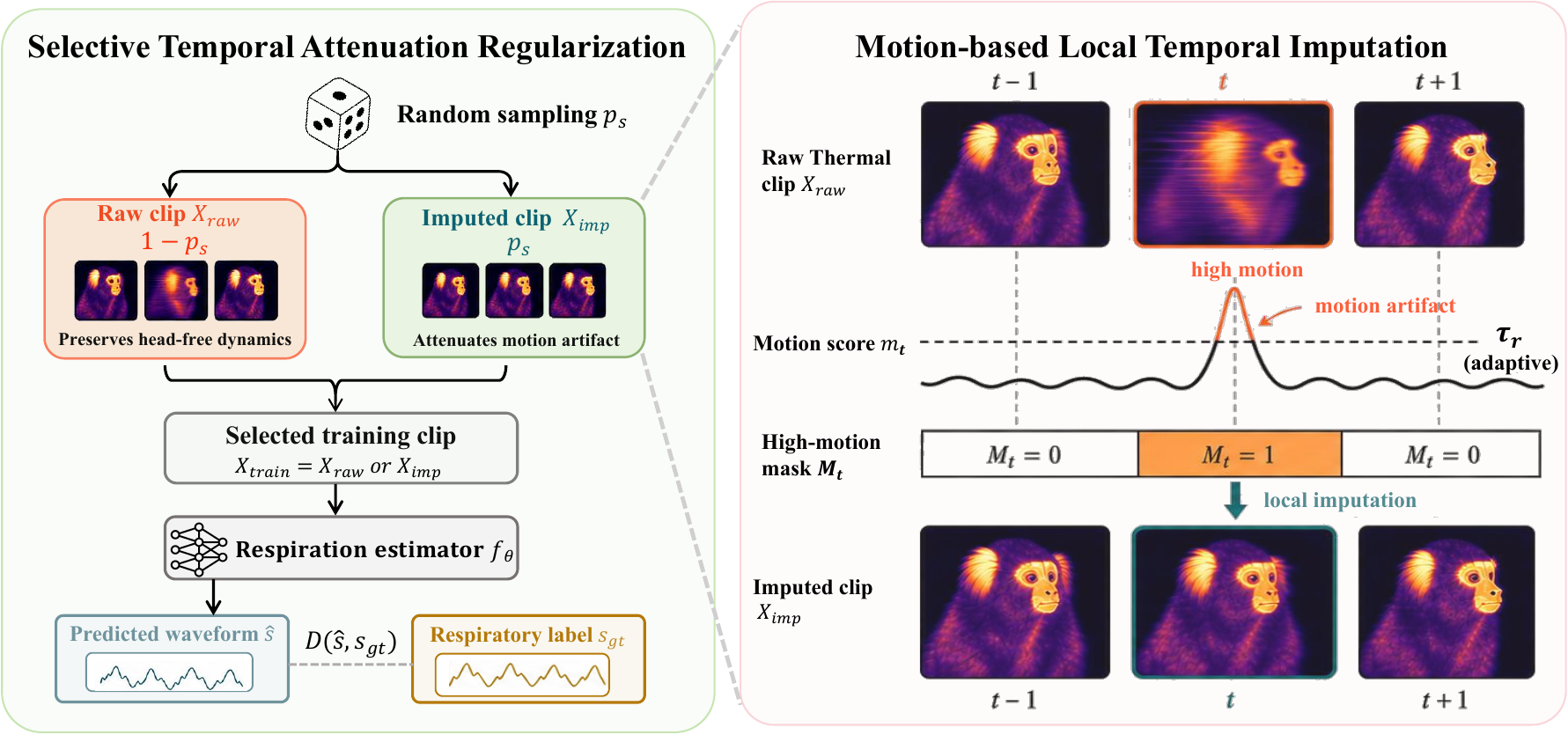}
    \caption{Overview of \method. A spatiotemporal network predicts
        a respiratory waveform from thermal clips. During training, high motion
        frames are identified from adjacent thermal differences and locally
        imputed using temporal neighbors with probability $p_s$, attenuating motion-induced thermal discontinuity while inference uses raw
        clips.}
    \label{fig:method}
\end{figure*}

\subsection{Thermal Respiratory Signal Recovery}

Existing video based physiological backbones learn spatiotemporal
representations from video under supervised learning.
These models recover weak physiological signals by exploiting temporal and
spatial dependencies in the input sequence, making them suitable for thermal
respiratory monitoring.
Instead of reducing a thermal clip to a single ROI averaged temperature trace,
we model the task as recovering a continuous respiratory signal from the full
spatiotemporal thermal sequence.

We adapt three representative backbones to thermal respiratory waveform
prediction: TS-CAN~\cite{liu2020multi}, PhysNet~\cite{PhysNet}, and
PhysFormer~\cite{yu2022physformer}.
Each model takes a thermal facial clip as input and predicts a continuous
respiratory signal.
This formulation keeps the frequency readout comparable to the traditional ROI
baseline while replacing the fragile ROI averaged trace with a learned waveform
extractor.

Given input clip $\mathbf{X}$, the model predicts a waveform
\begin{equation}
\hat{s}=f_{\theta}(\mathbf{X}).
\end{equation}
Respiratory rate is derived from the predicted waveform using the same
dominant-frequency readout as the ROI baseline.

\subsection{\methodfull}
Existing video based physiological models often improve robustness through model internal
mechanisms, such as attention modules,
difference frame inputs or temporal shift modules
~\cite{chen2018deepphys,liu2020multi,yu2022physformer}.
These mechanisms improve spatiotemporal feature extraction,
but they do not explicitly address motion-corrupted input observations during training.
In head-free marmoset thermal video, raw clips can contain short, high amplitude
motion induced thermal changes caused by head pose changes and changing nostril
orientation.
These changes are visually prominent, but they do not correspond to abrupt
changes in the contact respiration waveform.
Thus, the training data contains time points where the local video observation is
inconsistent with the physiological label.

\method addresses this input level mismatch without
modifying the backbone architecture.
It constructs a complementary training view by imputing selected high
motion frames from temporal neighbors while keeping the same respiratory label.
Stochastic exposure to both raw and attenuated clips encourages the model
to rely on respiration-consistent thermal dynamics, while inference remains
unchanged on raw thermal video.
Given a raw thermal clip $\mathbf{X}_{raw}=\{X_t\}_{t=1}^{T}$, STAR first
computes a framewise motion score from adjacent thermal differences:
\begin{equation}
d(t)=\mathrm{mean}\left(|X_t-X_{t-1}|\right), \qquad t=2,\ldots,T.
\end{equation}
Adjacent frame differences are used only to locate likely motion-corrupted frames, not as
a normalized difference input to the backbone.
High motion frames are selected adaptively within each clip by thresholding the
motion score at a percentile:
\begin{equation}
M_t=\mathbb{I}\left(d(t)\geq
\mathrm{Percentile}_{100(1-r_m)}
\left(\{d(t)\}_{t=2}^{T}\right)\right),
\end{equation}
where $r_m$ is the selected frame ratio.
We set $r_m=0.10$ by default, selecting approximately the top 10\% high motion
frames in each clip.

For a selected high motion frame, the current observation may contain an abrupt
thermal change that is not caused by respiration.
We assume that, in the absence of strong motion artifacts,
the underlying respiration related thermal observation varies smoothly over adjacent frames.
Let $Z_t$ denote the latent respiration-consistent thermal observation.
This assumption can be written as
\begin{equation}
Z_{t+1}-2Z_t+Z_{t-1}\approx 0,
\qquad
Z_t \approx \frac{1}{2}\left(Z_{t-1}+Z_{t+1}\right).
\end{equation}

Because $Z_t$ is not directly available, STAR uses the two neighboring observed
frames to form a symmetric local estimate for selected high motion frames:
\begin{equation}
\label{eq:star_imputation}
X_t^{imp}=
\begin{cases}
\frac{1}{2}\left(X_{t-1}+X_{t+1}\right),
& \text{if } M_t=1 \text{ and } 1<t<T,\\
X_t,
& \text{otherwise}.
\end{cases}
\end{equation}
Frames outside the high motion mask and boundary frames remain unchanged.
This operation preserves clip length and temporal alignment with the respiratory
waveform label.
It is not intended to reconstruct a visually perfect thermal frame; instead, it
selectively attenuates abrupt frame-level thermal discontinuities at likely
motion-corrupted time points while keeping low-motion frames raw.

During training, we use stochastic raw/attenuated exposure:
\begin{equation}
\mathbf{X}_{train}=
\begin{cases}
\mathbf{X}_{raw}, & \text{with probability } 1-p_s,\\
\mathbf{X}_{imp}, & \text{with probability } p_s.
\end{cases}
\end{equation}
The raw view preserves the input distribution used at inference,
while the attenuated view reduces local observation-label mismatch during training.
Inference always uses raw thermal clips.
% Appendix~\ref{sec:appendix_protocol_details} provides the step by step implementation.

\section{Experiments}
\label{sec:experiments}

We evaluate thermal respiratory signal recovery under progressively more
challenging conditions.
We first test whether thermal video contains recoverable respiratory information
under head-fixed conditions.
We then evaluate head-free respiratory estimation, where pose changes and
nostril orientation changes can corrupt local thermal evidence.
For learned models, we report both individual-dependent and
individual-independent evaluation, following common participant-dependent and
participant-independent protocols in video-based physiological measurement~\cite{chen2018deepphys}.
The tracked ROI frequency baseline from Section~\ref{sec:data} is included as a
non-learning reference.
% Finally, we ablate the main design choices of \method.

\paragraph{Protocol.}
All learned marmoset models use the same data split,
clip construction, respiratory rate estimation procedure, and evaluation code.
The main setting is individual-dependent cross-session evaluation.
In each split, one recording session is held out for testing,
while the remaining sessions are used for training and validation.
The same individual marmoset may appear across splits, but recording sessions are disjoint across
training, validation, and test sets.
This prevents window-level leakage and evaluates within-individual
generalization across recording sessions.
We additionally report four-fold individual-independent evaluation as a harder
setting.
In each fold, one individual marmoset is held out for testing,
and the remaining individuals are used for training and validation.
This setting tests cross-individual generalization under differences in thermal
appearance, facial geometry, and motion patterns.

Each input sample contains 50 thermal frames,
following short sequence spatiotemporal modeling protocols of ~\cite{PhysNet}.
For respiratory rate estimation, predictions from clips are aggregated into a 20\,s
analysis window following the frequency domain estimation protocol
of~\cite{cennini2010heart}.
The aggregated signal is filtered in the 0.8--2.5 Hz respiratory
band ~\cite{bishop2022marmosetbreathing,borjon2016arousal},
and respiratory rate is computed from the dominant spectral peak.
The contact respiration signal is processed with the same procedure to obtain the reference rate.

\paragraph{Training.}
All learned models use temporal signal supervision from the synchronized contact respiration signal.
For each backbone, we follow the corresponding training objective and
implementation practice in rPPG-Toolbox~\cite{liu2023rppg}.
Following standard rPPG preprocessing practice,
thermal frames are cropped to a fixed facial ROI before clip construction.
All models are trained with AdamW for 20 epochs, batch size 16, and learning rate
$3\times10^{-4}$.
For \method, we use $r_m=0.10$ and $p_s=0.50$ by default.

\paragraph{Metrics.}
We report MAE, MAPE, and Pearson correlation as defined in
Section~\ref{sec:data}.
Additional preprocessing and implementation details appear in
Appendix~\ref{sec:appendix_protocol_details}.

\subsection{Head-fixed Respiratory Signal Recovery}

The head-fixed experiment asks whether waveform prediction can reduce the
dependence on a manually defined ROI.
Table~\ref{tab:fixed_dl} compares the manual ROI frequency baseline with three
adapted video based physiological signal recovery backbones.
All learned models use the same head-fixed evaluation windows and
the same frequency domain respiratory rate estimation procedure.
TS-CAN provides a modest reduction in MAE and MAPE,
while PhysNet and PhysFormer substantially improve both error and correlation.
PhysFormer achieves the best overall performance,
with 0.82 bpm MAE and a Pearson correlation of 0.97.
These results show that learned spatiotemporal models can recover
fixed-head respiratory signals more reliably than a temperature trace obtained
by averaging a manually selected nasal ROI.

\begin{table}[!htbp]
    \caption{Head-fixed comparison between traditional ROI methods and
        spatiotemporal waveform prediction.}
    \label{tab:fixed_dl}
    \tablestyle{8pt}{1.08}
    \begin{tabular}{lccc}
        \toprule
        Method & MAE $\downarrow$ & MAPE $\downarrow$ & $\rho$ $\uparrow$ \\
        \midrule
        Manual ROI & 2.46 & 3.57 & 0.87 \\
        TS-CAN & 2.16 & 3.10 & 0.83 \\
        PhysNet & 0.93 & 1.40 & 0.95 \\
        PhysFormer & \textbf{0.82} & \textbf{1.21} & \textbf{0.97} \\
        \bottomrule
    \end{tabular}
\end{table}

\begin{figure}[t]
    \centering
    \begin{subfigure}[t]{0.42\linewidth}
        \centering
        \includegraphics[width=\linewidth]{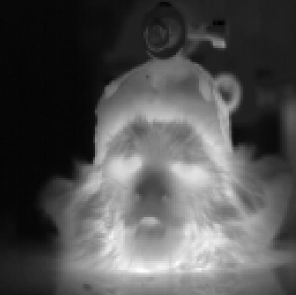}
        \caption{Thermal frame}
    \end{subfigure}
    \hfill
    \begin{subfigure}[t]{0.42\linewidth}
        \centering
        \includegraphics[width=\linewidth]{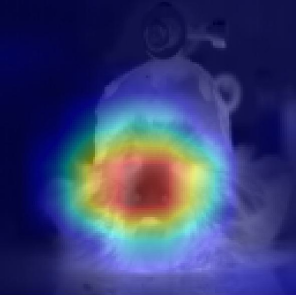}
        \caption{Grad-CAM}
    \end{subfigure}
    \caption{Grad-CAM visualization under head-fixed conditions.}
    \label{fig:gradcam}
\end{figure}

\paragraph{Interpretability}
We next examine whether the head-fixed model uses physiologically plausible spatial evidence.
We use Grad-CAM~\cite{selvaraju2017grad}, which visualizes image regions that
contribute strongly to the model output through gradient based localization.
Figure~\ref{fig:gradcam} shows Grad-CAM examples for thermal respiratory signal recovery.
The high response regions concentrate around the nostril and nearby local thermal structures,
rather than spreading uniformly across the full crop.
This pattern agrees with the ROI sensitivity analysis in Section~\ref{sec:data}:
the respiratory cue is local,
but the learned model can combine multiple local thermal regions instead of
collapsing the input to one fixed ROI average.

\subsection{Head-Free Robust Respiration Estimation}

The main experiment evaluates thermal respiratory signal recovery under head-free motion.
This setting is challenging because pose changes, nostril orientation changes,
fur, and partial occlusion can corrupt local thermal traces.

Table~\ref{tab:headfree_dl} compares head-free respiratory-rate estimation under
two evaluation settings.
For comparison, we also include the tracked CSRT ROI frequency baseline from Section~\ref{sec:data}.
This baseline is a non-learning method and does not involve train or test splits.
The upper block reports individual-dependent evaluation,
while the lower block reports individual-independent evaluation.

\paragraph{Individual dependent performance.}
Under individual dependent evaluation,
all learned signal recovery models substantially outperform the tracked ROI baseline.
Although CSRT achieves high visual tracking availability, its respiratory-rate
error remains extremely large (20.76 bpm MAE), indicating that tracking
continuity does not guarantee physiological signal fidelity.
This gap is important for head-free thermal sensing: the tracked box can remain
spatially stable even when the nostril heat exchange is weak, off-axis, or mixed
with fur and background temperature changes.
In contrast, learned models reduce the error by a large margin.
PhysFormer and PhysNet achieve the strongest performance, both reaching below 4 bpm MAE
and Pearson correlation above 0.90,
while TS-CAN remains less accurate but still outperforms ROI based tracking.
Moreover, \method consistently improves all evaluated backbones, suggesting that
motion-guided temporal attenuation mitigates the mismatch between motion-corrupted
frames and synchronized respiratory labels in the individual-dependent setting.

\paragraph{Individual independent performance.}
Under individual independent evaluation, performance degrades across all learned models.
This indicates that inter-individual domain shift is a major challenge for head-free respiratory estimation.
Learned respiratory representations may still depend on individual-specific thermal
appearance, facial geometry, and motion patterns.
This gap separates motion corruption within known individuals from generalization
to unseen individuals, which remains outside the direct objective of \method.
% Among the learned models, \method clearly improves PhysNet and provides limited gains for PhysFormer.
% Its effect on TS-CAN is mixed, slightly improving MAPE and correlation while not improving MAE. 
% These results suggest that \method mainly reduces motion-induced observation mismatch, but does not fully resolve cross-individual domain shift.
Among the learned models, \method clearly improves PhysNet and gives limited
gains for PhysFormer.
For TS-CAN, it slightly improves MAPE and correlation but not MAE.
These results suggest that \method reduces motion-induced observation mismatch,
but does not fully resolve cross-individual domain shift.

\begin{table}[ht]
    \caption{Head-free respiratory rate estimation under individual-dependent
        and individual-independent evaluation.}
    \label{tab:headfree_dl}
    \tablestyle{8pt}{1.1}
    \begin{tabular}{lccc}
        \toprule
        Method & MAE $\downarrow$ & MAPE $\downarrow$ & $\rho$ $\uparrow$ \\
        \midrule
        \multicolumn{4}{l}{\textit{Tracked ROI reference}} \\
        CSRT ROI & 20.76 & 21.55 & 0.10 \\
        \midrule
        \multicolumn{4}{l}{\textit{Individual dependent}} \\
        PhysFormer & 3.31 & 3.98 & 0.91 \\
        PhysFormer + \method & 2.71 & 3.28 & 0.94 \\
        PhysNet & 2.69 & 3.54 & 0.92 \\
        PhysNet + \method & \textbf{2.23} & \textbf{2.90} & \textbf{0.94} \\
        TS-CAN & 7.97 & 9.81 & 0.64 \\
        TS-CAN + \method & 7.05 & 9.11 & 0.70 \\
        \midrule
        \multicolumn{4}{l}{\textit{Individual independent}} \\
        PhysFormer & 9.76 & 11.69 & 0.53 \\
        PhysFormer + \method & 9.49 & 11.68 & 0.58 \\
        PhysNet & 8.56 & 9.93 & 0.59 \\
        PhysNet + \method & \textbf{7.06} & \textbf{8.78} & \textbf{0.67} \\
        TS-CAN & 11.70 & 14.69 & 0.48 \\
        TS-CAN + \method & 11.81 & 14.61 & 0.51 \\
        \bottomrule
    \end{tabular}
\end{table}

Additional motion-stratified comparisons between learned waveform prediction and
ROI frequency analysis are provided in
Appendix~\ref{sec:appendix_motion_details}.

\subsection{Ablation Study}
\paragraph{Ablation protocol.}
We conduct ablations on the B6 head-free subset using an individual-dependent protocol, where training, validation, and test samples are split
by recording session.
This controlled setting isolates the effect of \method design choices from
inter-individual variation.

\paragraph{Temporal operation.}
This ablation isolates three factors in \method: motion guided frame selection,
the choice between smoothing and imputation for selected high motion frames,
and whether the gain can be explained by replacing frames with lower motion observations.
The raw baseline applies no temporal operation.
The proposed imputation follows Eq.~\eqref{eq:star_imputation}:
selected high motion frames are replaced by an adjacent frame estimate,
while other frames remain unchanged.
As a comparison, temporal filtering applies a centered three frame kernel,
\[ I_t^{\mathrm{filter}} = 0.25 I_{t-1} + 0.5 I_t + 0.25 I_{t+1}. \]
This filter is equivalent to a convex combination of the current frame and
the adjacent frame estimate used by \method. Therefore, when the current frame is motion corrupted,
filtering can still retain part of the corrupted observation.
Random filtering and random imputation apply temporal operations at randomly selected time points,
testing whether temporal modification alone is sufficient without motion guided selection.
Low motion imputation replaces selected high motion frames with randomly sampled low motion frames from the same session,
testing whether lower motion appearance alone explains the gain.

The ablation results in Table~\ref{tab:star_ablation} support this design.
Random filtering and random imputation do not improve the raw baseline,
indicating that temporal operations must be targeted to motion-corrupted frames.
% Low-motion imputation performs worse, suggesting that replacing stable frames can
% remove useful respiratory information.
Low-motion imputation performs worse, indicating that lower motion is insufficient 
without temporal alignment to the respiratory phase.
Temporal filtering degrades performance, consistent with the fact that it
still retains part of the corrupted center frame.
The best result is obtained only when the operation is both motion-guided and based
on local temporal imputation.
This supports viewing \method as a targeted reliability intervention rather than
generic temporal augmentation: it modifies frames only when motion makes the
local thermal evidence less trustworthy.

\begin{table}[!htbp]
    \caption{STAR ablation on B6 under individual-dependent evaluation.}
    \label{tab:star_ablation}
    \tablestyle{8pt}{1.1}
    \begin{tabular}{lccc}
        \toprule
        Variant & MAE $\downarrow$ & MAPE $\downarrow$ & $\rho$ $\uparrow$ \\
        \midrule
        \multicolumn{4}{l}{\textit{Operation}} \\
        Raw & 1.13 & 1.32 & 0.98 \\
        Random filter & 1.19 & 1.36 & 0.97 \\
        Low-motion & 2.09 & 2.45 & 0.93 \\
        Filter & 2.00 & 2.39 & 0.94 \\
        Random imput. & 1.26 & 1.53 & 0.95 \\
        STAR & \textbf{0.97} & \textbf{1.11} & \textbf{0.98} \\
        \midrule
        \multicolumn{4}{l}{\textit{Raw/attenuated probability}} \\
        Filter, $p_s=0.50$ & 2.00 & 2.39 & 0.94 \\
        Filter, $p_s=1.00$ & 4.42 & 4.91 & 0.77 \\
        STAR, $p_s=0.50$ & \textbf{0.97} & \textbf{1.11} & \textbf{0.98} \\
        STAR, $p_s=1.00$ & 1.22 & 1.43 & 0.97 \\
        \midrule
        \multicolumn{4}{l}{\textit{Mask ratio}} \\
        Top 5\% & 1.75 & 2.11 & 0.94 \\
        Top 10\% & \textbf{0.97} & \textbf{1.11} & \textbf{0.98} \\
        Top 20\% & 1.87 & 2.23 & 0.94 \\
        \bottomrule
    \end{tabular}
\end{table}

% \paragraph{Raw/attenuated exposure probability and motion mask ratio.}
\paragraph{Exposure probability and mask ratio.}
We further ablate two parameters of \method: the probability $p_s$ of using the
attenuated view during training, and the fraction of frames selected by the motion
mask. As shown in Table~\ref{tab:star_ablation}, stochastic raw and attenuated exposure ($p_s=0.50$) gives the best performance.
For \method, increasing $p_s$ to 1.00 causes only a moderate drop,
suggesting that local temporal imputation produces an attenuated view that remains
largely consistent with respiratory signal recovery.
In contrast, temporal filtering degrades sharply when always applied ($p_s=1.00$),
indicating that generic filtering introduces a stronger train test mismatch and
can suppress useful respiratory dynamics while retaining part of the corrupted center frame.

The motion mask ratio controls how many high-motion frames are selected for local imputation.
Selecting too few frames may miss motion artifacts, whereas selecting too many
frames may overprocess frames that still contain useful respiratory dynamics.
The top 10\% setting gives the best balance.
Together, these results support the default setting of $p_s=0.50$ and a top
10\% motion mask.
This behavior matches the goal of \method: correcting likely motion corruption
while leaving most respiratory evidence unchanged.

\section{Discussion and Conclusion}
\label{sec:conclusion}

In head-free marmosets, visually plausible nasal tracking can fail to preserve
respiratory signals, motivating learned waveform recovery from thermal video.
\method applies local temporal imputation to selected high-motion frames during
training to reduce motion-induced observation-label mismatch.
These findings may also inform learning from perturbed visual
evidence~\cite{tong2022videomae,zhong2020random} and sensing affected by motion
artifacts or cross-sensor misalignment~\cite{zaitsev2015motion,huang2022multimodal}.

Our modest cohort and controlled recordings limit generalization, particularly
across individuals.
Future work should expand the cohort and develop quality control for naturalistic
motion, occlusion, camera distance changes, and social interactions.

\section*{Acknowledgment}
This work was supported by the Strategic Priority Research Program of the Chinese
Academy of Sciences under Grant XDB1010102, the STI 2030---Major Projects under
Grant 2021ZD0201500, and the National Natural Science Foundation of China under
Grant T2341003.

\section*{Compliance with Ethical Standards}
All animal procedures were conducted in accordance with institutional and national
animal care guidelines and were approved by the Institutional Animal Care and Use
Committee (IACUC) of Tsinghua University.

{
    \small
    \bibliographystyle{ieeenat_fullname}
    \bibliography{main}
}
\clearpage
\appendix

\section{Data Collection and Screening Details}
\label{sec:appendix_data_details}

\paragraph{Acquisition hardware.}
The acquisition system records thermal video and contact respiration on a shared
hardware time base.
Thermal video is captured using a MAG62 thermal imaging camera
(Magnity Electronics Co., Ltd.) with $640\times512$ spatial resolution, 50 Hz
frame rate, approximately 50 mK thermal sensitivity, and external trigger
support.
External triggering aligns each thermal frame with the contact respiration
reference.

The contact respiration reference is recorded using a Respiration Pillow Sensor
9'' (SAII), a noninvasive air bladder sensor for animal respiration monitoring.
During breathing, abdominal motion induces small pressure changes inside the
sealed bladder.
These pressure changes are transmitted to the sensing unit and converted into a
continuous analog voltage waveform.

During acquisition, the bladder is placed near the abdomen and secured with a
medical elastic tubular bandage.
The bandage maintains stable contact while avoiding excessive compression,
allowing the sensor to measure respiratory motion without contacting the nasal
or oral region.

\begin{figure}[!htbp]
    \centering
    \begin{subfigure}[t]{0.48\linewidth}
        \centering
        \includegraphics[width=0.78\linewidth]{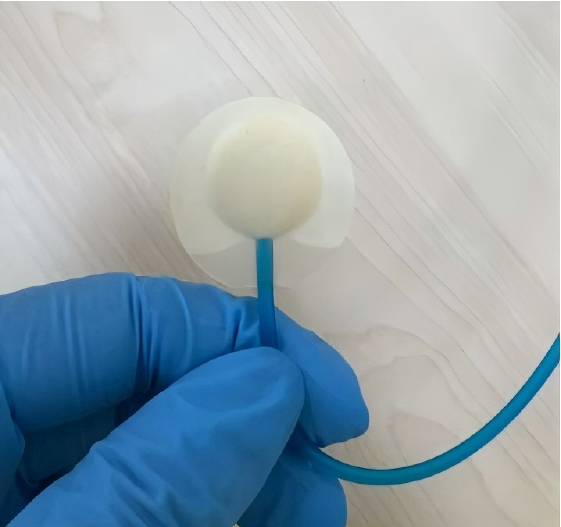}
        \caption{Respiration pillow sensor}
    \end{subfigure}
    \hfill
    \begin{subfigure}[t]{0.48\linewidth}
        \centering
        \includegraphics[width=0.78\linewidth]{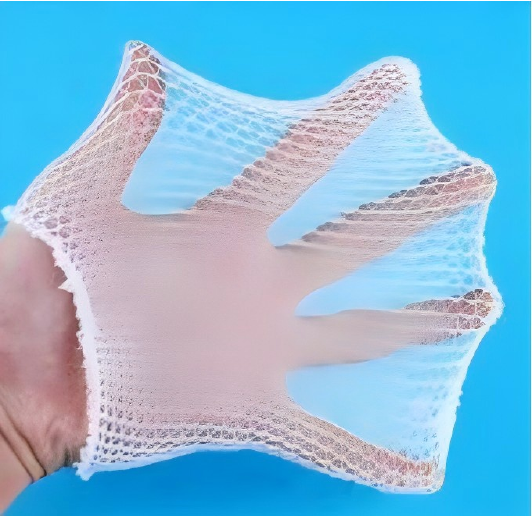}
        \caption{Elastic tubular bandage}
    \end{subfigure}
    \caption{Contact respiration reference hardware. The air bladder sensor
    converts abdominal respiratory motion into a pressure signal, while the
    elastic tubular bandage maintains stable contact during acquisition.}
    \label{fig:appendix_respiration_sensor}
\end{figure}

\begin{figure}[!htbp]
    \centering
    \begin{subfigure}[t]{0.48\linewidth}
        \centering
        \includegraphics[width=\linewidth,height=1.75in,keepaspectratio]{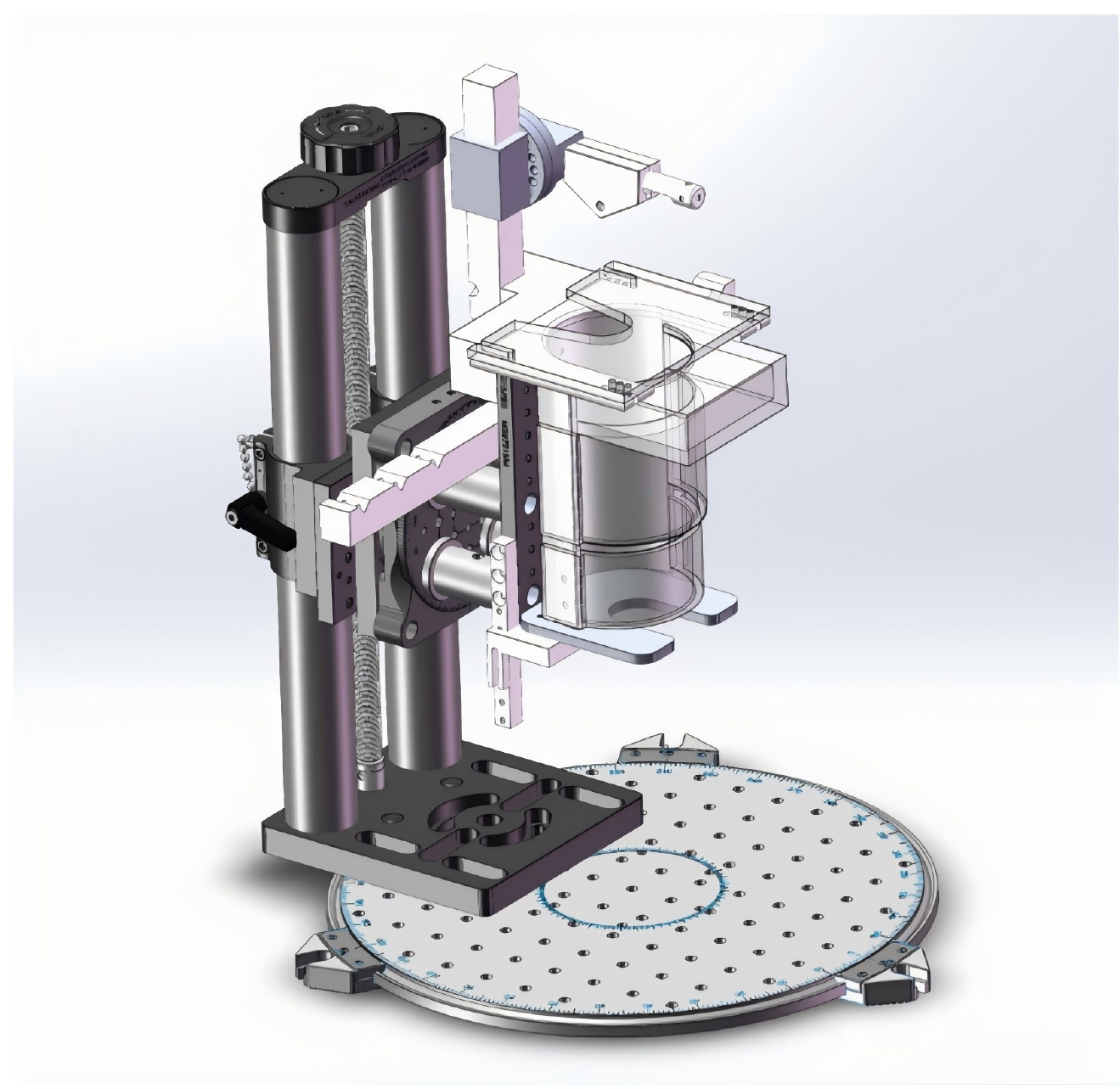}
        \caption{Restraint apparatus}
        \label{fig:appendix_restraint}
    \end{subfigure}
    \hfill
    \begin{subfigure}[t]{0.48\linewidth}
        \centering
        \includegraphics[width=\linewidth,height=1.75in,keepaspectratio]{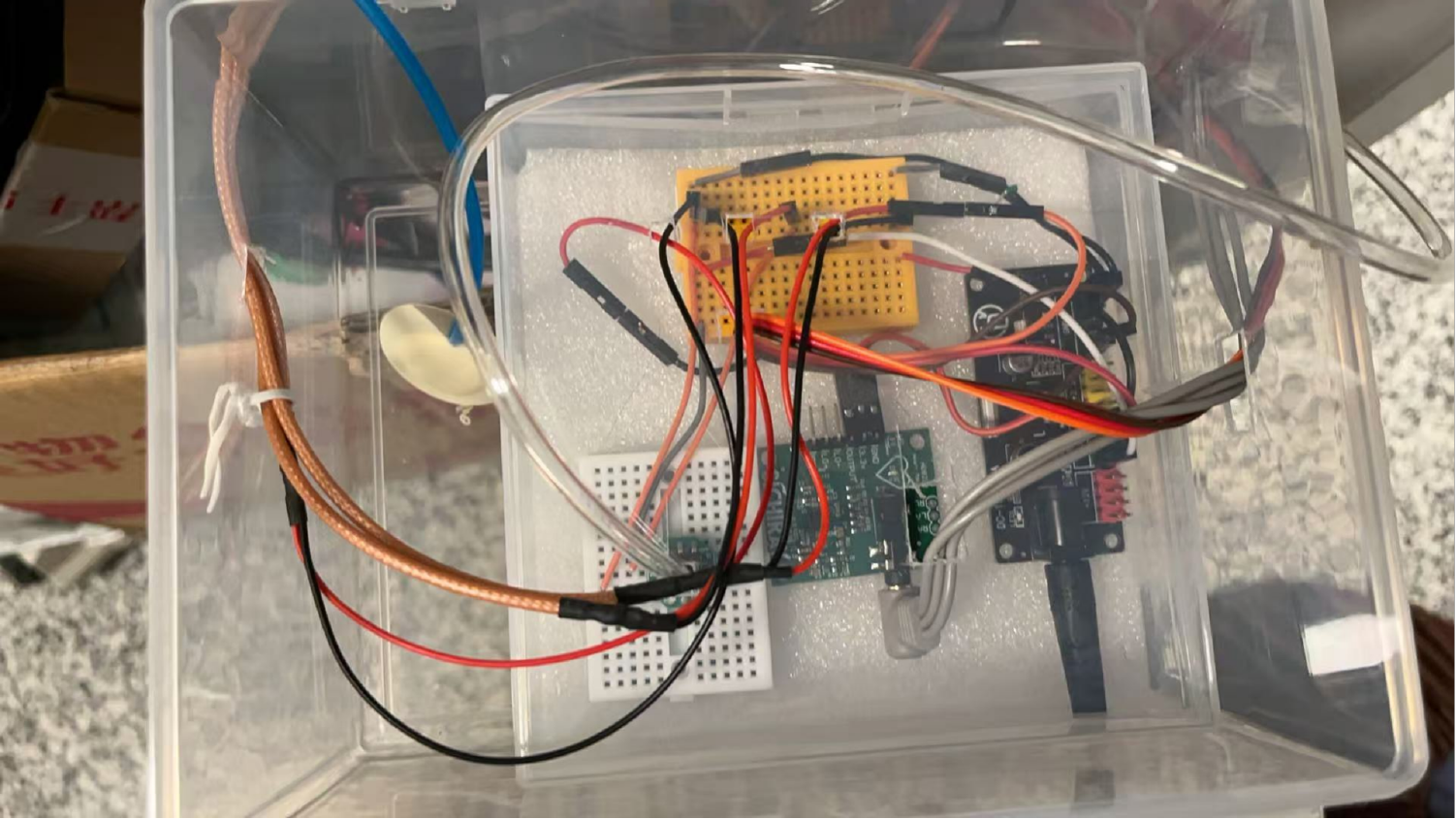}
        \caption{Sensor components}
        \label{fig:appendix_sensor_components}
    \end{subfigure}
    \caption{Synchronized marmoset thermal respiration acquisition setup.}
    \label{fig:appendix_acquisition_setup}
\end{figure}

\paragraph{Subjects and sessions.}
The full collection contains 48 five minute sessions, corresponding to
240 minutes of synchronized thermal video and contact respiration from four male
marmosets.
The subjects are B4, B6, 200106, and ZJM, with approximate body weights of 438.7
g, 436.7 g, 433.2 g, and 395.7 g, respectively.
Each recording session lasts 5 minutes, which balances thermal camera
stabilization against behavioral disturbance caused by long restraint.
The head-free condition contains 32 sessions (160 minutes), and the fixed-head
condition contains 16 sessions (80 minutes).
B4 and B6 were recorded under head-free conditions, while 200106 and ZJM were
recorded under both fixed-head and head-free conditions.

\begin{figure}[!htbp]
    \centering
    \begin{minipage}{0.98\columnwidth}
        \centering
        \begin{subfigure}[t]{\linewidth}
            \centering
            \includegraphics[width=\linewidth]{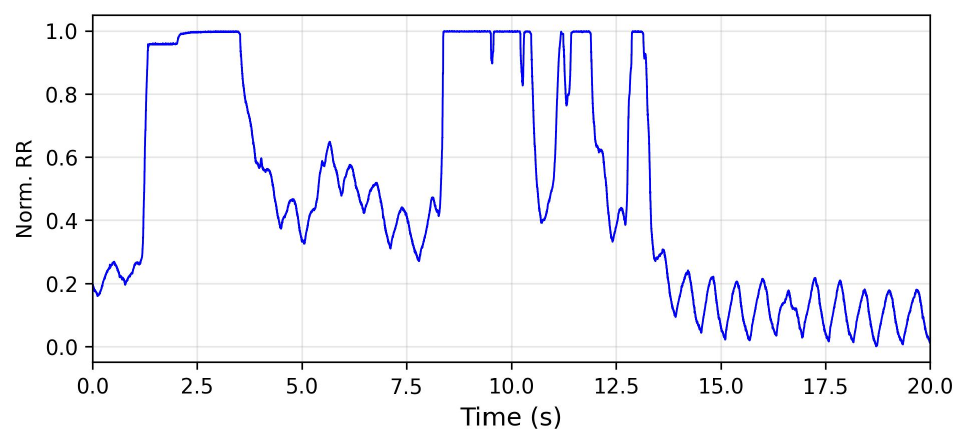}
            \caption{Excluded abnormal window}
        \end{subfigure}
        \vspace{0.5em}
        \begin{subfigure}[t]{\linewidth}
            \centering
            \includegraphics[width=\linewidth]{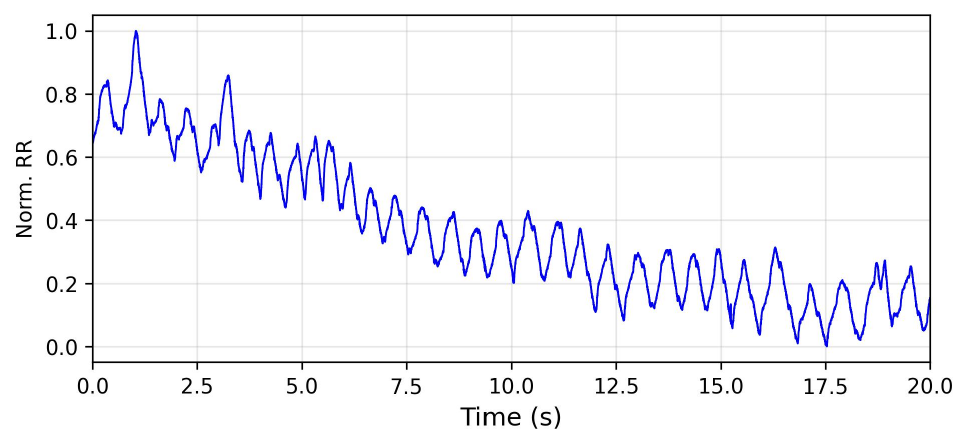}
            \caption{Retained borderline window}
        \end{subfigure}
        \caption{Examples of respiratory signal screening.}
        \label{fig:appendix_rr_screening_examples}
    \end{minipage}
\end{figure}

\paragraph{Respiratory signal screening examples.}
Figure~\ref{fig:appendix_rr_screening_examples} shows representative examples
from respiratory signal screening.
This screening is applied to the contact reference signal before respiratory
rate evaluation.
Windows with severely corrupted respiratory signals are excluded because the
reference rate is unreliable.
Borderline windows are retained when motion causes waveform amplitude changes but
the respiratory periodicity remains distinguishable.

\FloatBarrier

\section{Evaluation and Implementation Details}
\label{sec:appendix_protocol_details}

\paragraph{Post hoc oracle ROI search.}
The oracle ROI search supports the fixed-head ROI sensitivity analysis in
Section~\ref{sec:data}.
It is used only as a diagnostic analysis to test whether the error of a practical
manual ROI is caused by weak thermal evidence or imperfect ROI placement.

Because head size, camera distance, and manual ROI scale vary across recordings,
we normalize the search space by nostril distance (ND).
The left and right nostril annotations define the nostril center $(c_x,c_y)$ and
the distance $d_{\mathrm{ND}}$.
A candidate ROI is specified by a normalized center offset and normalized width
and height:
\begin{equation}
\begin{aligned}
\Omega(s_x,s_y,q_w,q_h)
&=
\mathrm{Box}(
c_x+s_x d_{\mathrm{ND}},
c_y+s_y d_{\mathrm{ND}},\\
&\qquad
q_w d_{\mathrm{ND}},
q_h d_{\mathrm{ND}}).
\end{aligned}
\end{equation}
Candidate boxes that extend outside the thermal frame are clipped to the valid
image region.

The search uses a coarse to fine procedure.
In the coarse stage, the center offsets are
$s_x,s_y\in\{-1.0,-0.5,0,0.5,1.0\}$, and the width and height scales are
$q_w,q_h\in\{0.6,0.8,1.0,1.5,2.0,2.5,3.0\}$.
These candidates cover both small nostril adjacent regions and broader nasal
regions.
In the fine stage, we search around the best coarse candidate using center and
scale perturbations of
$\{-0.30,-0.15,0,0.15,0.30\}d_{\mathrm{ND}}$.

For every candidate ROI, we run the same frequency domain estimation pipeline
used in the main evaluation.
The ROI averaged thermal trace is detrended, normalized, filtered in the
respiratory band, and converted to respiratory rate by the dominant spectral
frequency.
The oracle ROI is selected by minimizing respiratory rate error over retained
fixed-head analysis windows:
\begin{equation}
\Omega_{\mathrm{oracle}}
=
\arg\min_{\Omega}
\frac{1}{N}\sum_{i=1}^{N}|\hat{r}_i(\Omega)-r_i|.
\end{equation}
This selection uses the contact reference error and is therefore not deployable.
It is used only as an oracle upper bound for ROI frequency analysis.

To further examine scale sensitivity, we group candidate ROIs by ND normalized
size and record the best result for each size.
Table~\ref{tab:appendix_roi_scale_sensitivity} shows that smaller local ROIs
near the nostril region provide stronger respiratory frequency estimates,
whereas larger ROIs dilute the respiratory signal with non respiratory pixels.

\begin{table}[!htbp]
    \caption{ND normalized ROI scale sensitivity under fixed-head conditions.}
    \label{tab:appendix_roi_scale_sensitivity}
    \tablestyle{8pt}{1.08}
    \begin{tabular}{@{}lccc@{}}
        \toprule
        Size / ND & MAE $\downarrow$ & MAPE $\downarrow$ & $\rho$ $\uparrow$ \\
        \midrule
        0.6 & \textbf{0.59} & \textbf{0.83\%} & \textbf{0.99} \\
        0.8 & 0.71 & 1.00\% & 0.98 \\
        1.0 & 0.79 & 1.12\% & 0.98 \\
        1.5 & 1.75 & 2.51\% & 0.92 \\
        2.0 & 1.80 & 2.65\% & 0.92 \\
        2.5 & 2.20 & 3.16\% & 0.89 \\
        3.0 & 2.54 & 3.71\% & 0.85 \\
        \bottomrule
    \end{tabular}
\end{table}

\paragraph{RR power map and ROI scale sensitivity.}
The RR power map in Figure~\ref{fig:best_roi} visualizes where respiratory-band
thermal energy is concentrated.
For each candidate ROI, we extract the mean thermal signal across pixels and
compute its PSD within each analysis window.
The signal is filtered to the respiratory range, normalized within the window,
and converted to the frequency domain using a periodogram.
We then integrate the PSD in a narrow band centered at the ground truth
respiratory-rate frequency.
This produces an RR-band power score for each candidate ROI.
Mapping these scores back to the ROI locations yields the RR power map.
Brighter regions indicate ROIs whose thermal fluctuations contain stronger
respiratory frequency power.

We also quantify ND normalized scale sensitivity.
For each ROI size, we search over candidate positions and record the best
frequency estimation result at that scale.

Table~\ref{tab:appendix_roi_scale_sensitivity} shows a clear trend.
Smaller ROIs provide stronger respiratory estimates.
The best scale in the searched range is 0.6 ND, with 0.59 bpm MAE,
0.83\% MAPE, and Pearson $\rho=0.99$.
At 3.0 ND, MAE increases to 2.54 bpm and Pearson $\rho$ decreases to 0.85.
This result supports the fixed-head conclusion: the marmoset thermal respiration
cue is concentrated in a small nostril adjacent region, and broader ROIs dilute
the periodic signal with non respiratory pixels.

\paragraph{ROI tracking and verified subset.}
\label{sec:appendix_tracking_details}
The tracking analysis in Section~\ref{sec:data} separates visual continuity from
physiological validity.
For head-free tracking experiments, KCF and CSRT are initialized from manually
annotated respiratory ROIs and updated frame by frame.
ROI availability measures whether the tracker returns a valid box, not whether
the box contains a usable respiratory signal.
It is defined as the fraction of frames in a 20 s window for which tracking
succeeds:
\begin{equation}
\mathrm{ROI\ availability}
=
\frac{N_{\mathrm{successful\ tracking\ frames}}}
{N_{\mathrm{total\ frames}}}.
\end{equation}
In implementation, a frame is successful when the OpenCV tracker returns
\texttt{ok = 1} and the predicted bounding box lies within the valid image
region.
A window is counted as visually available when ROI availability is at least
95\%.
For the main tracking analysis, we use a 1$\times$2ND bilateral nostril ROI
because it covers both nostrils under small pose changes.
We then manually inspect CSRT outputs and retain 141 windows in which the ROI
consistently covers the nasal or perinasal region.
This verified subset is used to show that visually plausible tracking still does
not guarantee accurate respiratory-rate estimation.

\paragraph{Training details.}
All learned marmoset models use temporal signal supervision from the synchronized
contact respiration signal.
For each backbone, we follow the corresponding training objective, validation
criterion, and implementation practice in rPPG-Toolbox
~\cite{liu2023rppg}.
Thermal frames are cropped to a fixed facial ROI before clip construction,
following common video based physiological measurement preprocessing.

Unless otherwise specified, models are trained with AdamW for 20 epochs, batch
size 16, and learning rate $3\times10^{-4}$.
All learned backbones use the same preprocessing, ROI crop, clip construction,
respiratory rate estimation procedure, and evaluation code.

For individual-dependent cross-session evaluation, recording sessions are
split so that training, validation, and test sets contain nonoverlapping
sessions.
The same individual marmoset may appear across splits, but clips from the same recording
session never appear in more than one split.
This prevents leakage at the clip or window level while evaluating within-individual
generalization across recording sessions.

Synchronized videos are segmented into fixed length clips before training and
evaluation.
Each input clip contains 50 frames, corresponding to 2\,s at the 25 fps training
sampling rate, and the model output length is 50.
Respiratory rate metrics are computed over 20\,s analysis windows.

\begin{table}[!htbp]
    \caption{Default training and evaluation configuration for the marmoset
        experiments.}
    \label{tab:appendix_training_config}
    \tablestyle{4pt}{1.08}
    \begin{tabular}{p{0.31\linewidth}p{0.62\linewidth}}
        \toprule
        Config & Value \\
        \midrule
        Backbones & TS-CAN, PhysNet, PhysFormer \\
        Input modality & Cropped thermal facial clip \\
        Training objective & Backbone specific objective following rPPG-Toolbox \\
        Clip length & 2\,s / 50 frames \\
        Sampling rate & 25 fps \\
        Input size & $1\times50\times128\times128$ \\
        Output length & 50 \\
        Analysis window & 20\,s / 500 frames \\
        Batch size & 16 \\
        Epochs & 20 \\
        Learning rate & $3\times10^{-4}$ \\
        Optimizer & AdamW \\
        Motion mask percentile & 0.90 \\
        Imputed sample probability $p_s$ & 0.50 \\
        Respiratory band & 0.80--2.50 Hz \\
        \bottomrule
    \end{tabular}
\end{table}

\section{Supplementary Agreement Analysis}
\label{sec:appendix_fixed_agreement}

\begin{figure}[!htbp]
    \centering
    \begin{subfigure}[t]{0.49\linewidth}
        \centering
        \includegraphics[width=\linewidth]{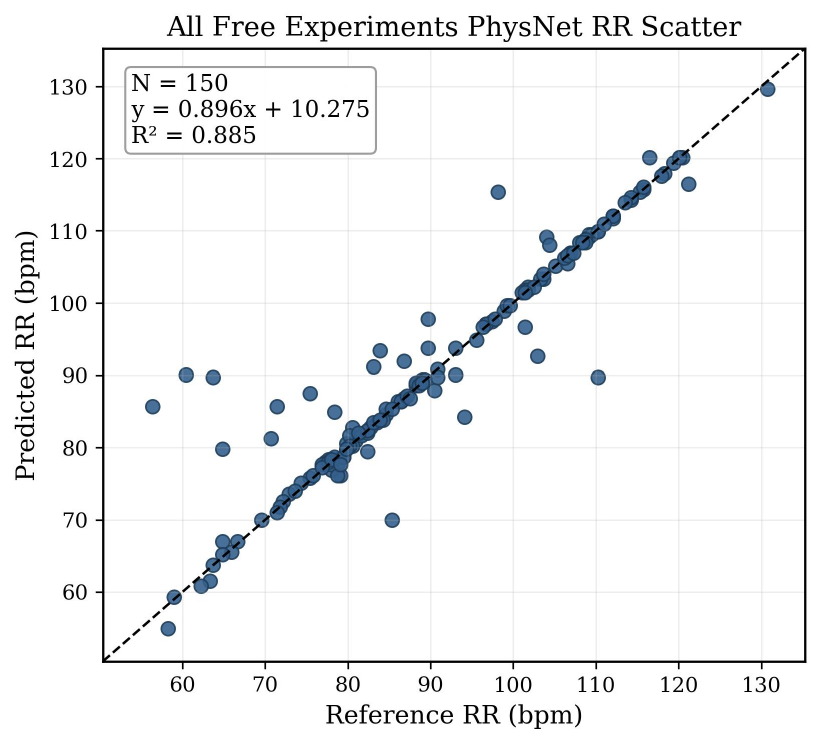}
        \caption{Scatter plot}
    \end{subfigure}
    \hfill
    \begin{subfigure}[t]{0.49\linewidth}
        \centering
        \includegraphics[width=\linewidth]{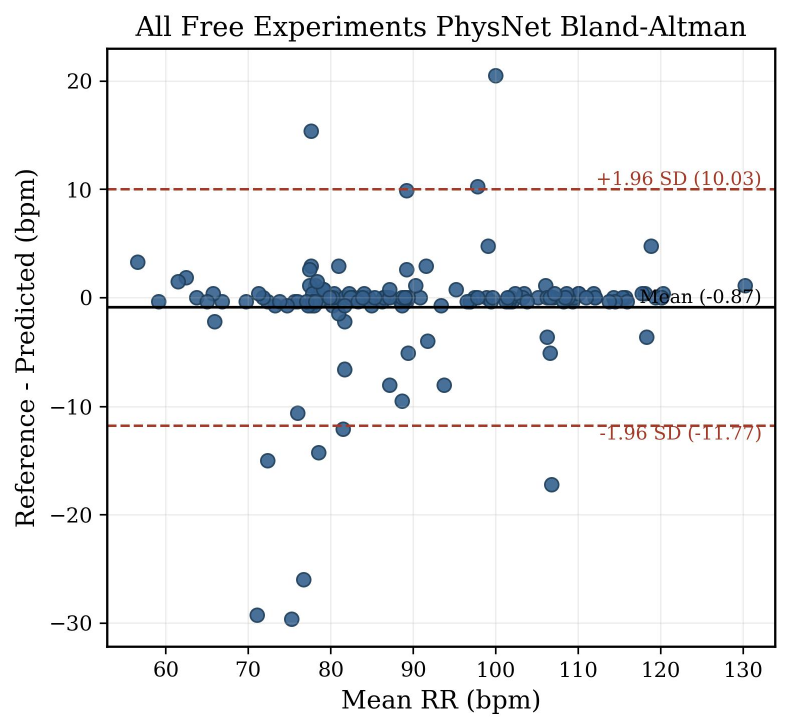}
        \caption{Bland--Altman plot}
    \end{subfigure}
    \caption{Supplementary agreement analysis for \method with PhysNet.}
    \label{fig:appendix_star_physnet_agreement}
\end{figure}

Figure~\ref{fig:appendix_star_physnet_agreement} provides an additional
agreement view for the STAR trained PhysNet model.
The scatter plot compares predicted and reference respiratory rates across
analysis windows.
The Bland--Altman plot shows the signed error as a function of the average
respiratory rate.
Together, these plots complement the MAE, MAPE, and correlation metrics in the
main text by showing the distribution of prediction errors.

\FloatBarrier

\section{Motion Stratification Analysis}
\label{sec:appendix_motion_details}

The main head-free results aggregate windows with different motion levels.
Here, we stratify the head-free windows by two complementary motion summaries:
motion intensity and motion duration.
Motion intensity captures short bursts of large thermal change, whereas motion
duration measures how much of a window is affected by motion.

For motion intensity, we compute the adjacent frame difference sequence
\begin{equation}
d(t)=\mathrm{mean}\left(|X_t-X_{t-1}|\right).
\end{equation}
The 95th percentile of $d(t)$ within each window is used as the motion intensity
score.

For motion duration, we threshold the framewise motion sequence and measure the
fraction of frames in the window that exceed the motion threshold.

\begin{figure}[!htbp]
    \centering
    \includegraphics[width=\linewidth]{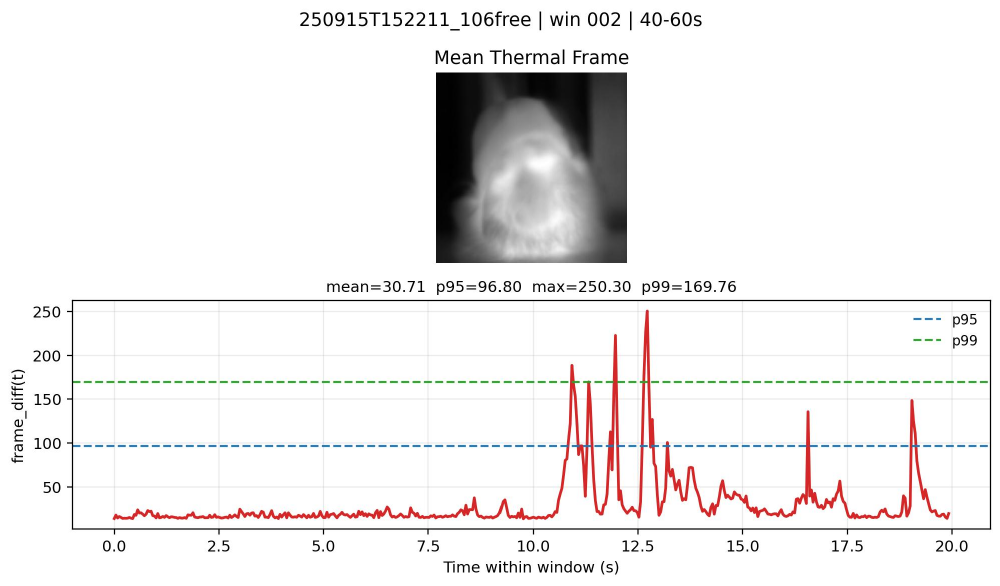}
    \caption{Motion intensity definition.}
    \label{fig:motion_intensity_definition}
\end{figure}

\begin{figure}[!htbp]
    \centering
    \includegraphics[width=\linewidth]{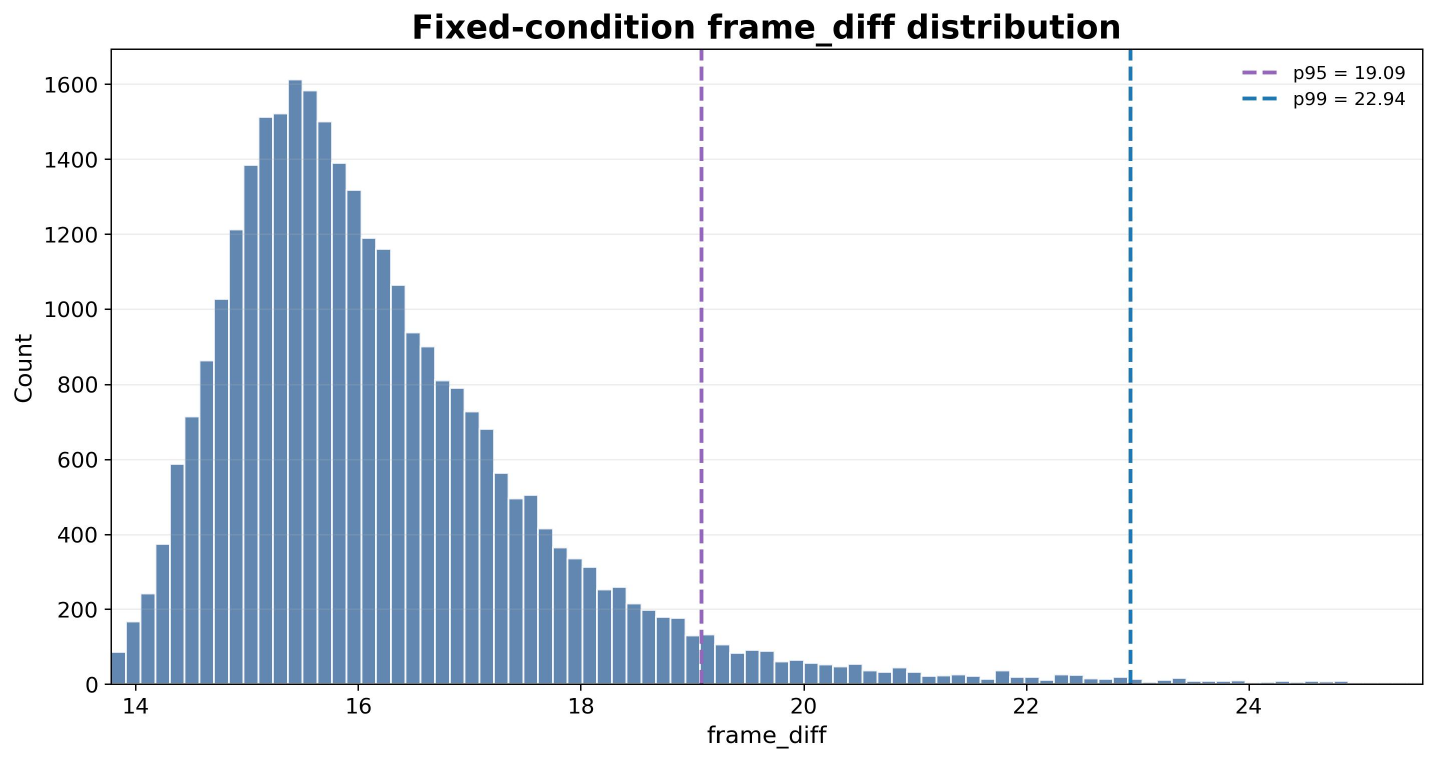}
    \caption{Motion duration definition.}
    \label{fig:motion_duration_definition}
\end{figure}

We compare ROI frequency analysis with learned respiratory signal recovery within
these motion strata.
Under motion intensity grouping, ROI frequency analysis degrades rapidly as
short burst thermal motion increases.
The learned model is also affected by motion, but maintains higher success rates,
especially under the 3 bpm and 5 bpm thresholds.

\begin{figure}[!htbp]
    \centering
    \includegraphics[width=\linewidth]{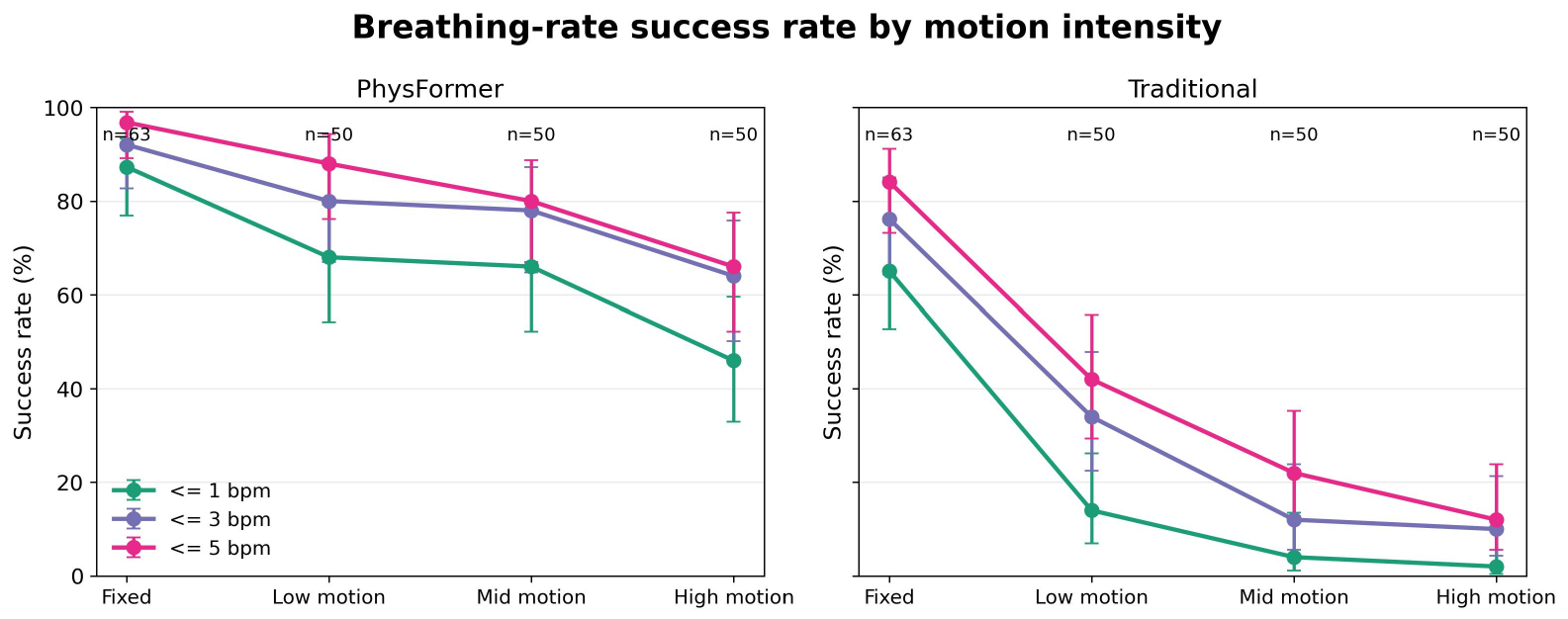}
    \caption{Motion intensity stratified success rates on head-free windows.}
    \label{fig:motion_intensity}
\end{figure}

Under motion duration grouping, ROI frequency analysis also degrades as the
fraction of motion affected frames increases.
The learned model maintains higher success rates across duration bins.

\begin{figure}[!htbp]
    \centering
    \includegraphics[width=\linewidth]{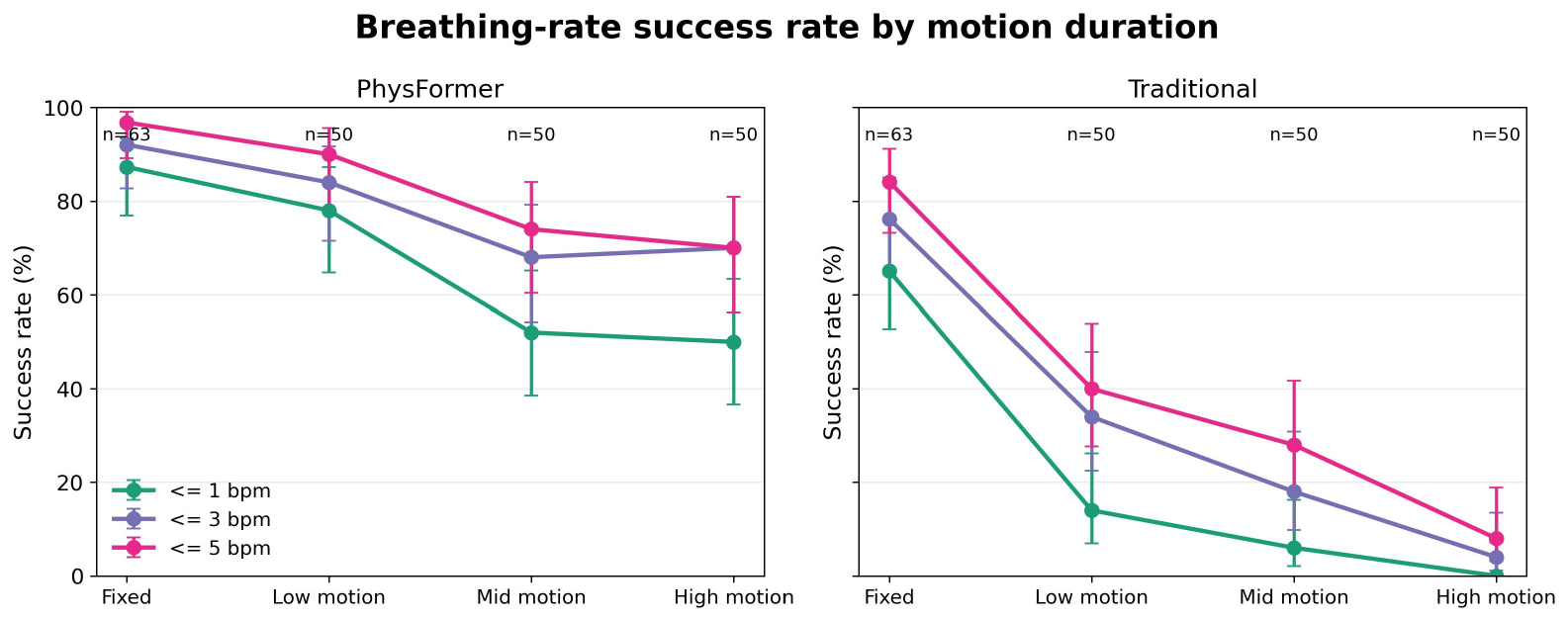}
    \caption{Motion duration stratified success rates on head-free windows.}
    \label{fig:motion_duration}
\end{figure}
\FloatBarrier

\section{Additional Tracking Examples}
\label{sec:appendix_tracking_examples}

These examples complement the tracking analysis in Section~\ref{sec:data}.
They are included to make the failure mode visually inspectable.
KCF and CSRT can differ in trajectory stability, and CSRT can maintain a
plausible nasal box even when the ROI averaged signal no longer tracks
respiration.

\begin{figure*}[!htbp]
    \centering
    \begin{subfigure}[t]{\linewidth}
        \centering
        \includegraphics[width=\linewidth]{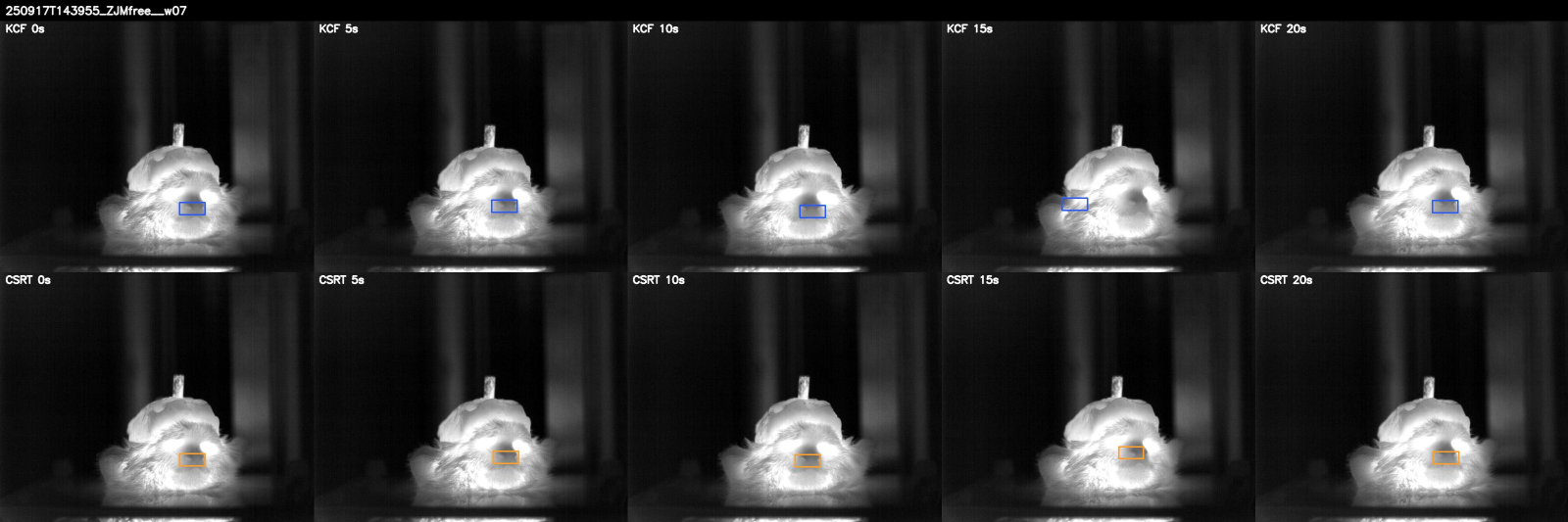}
        \caption{Example 1}
    \end{subfigure}
    \vspace{0.6em}

    \begin{subfigure}[t]{\linewidth}
        \centering
        \includegraphics[width=\linewidth]{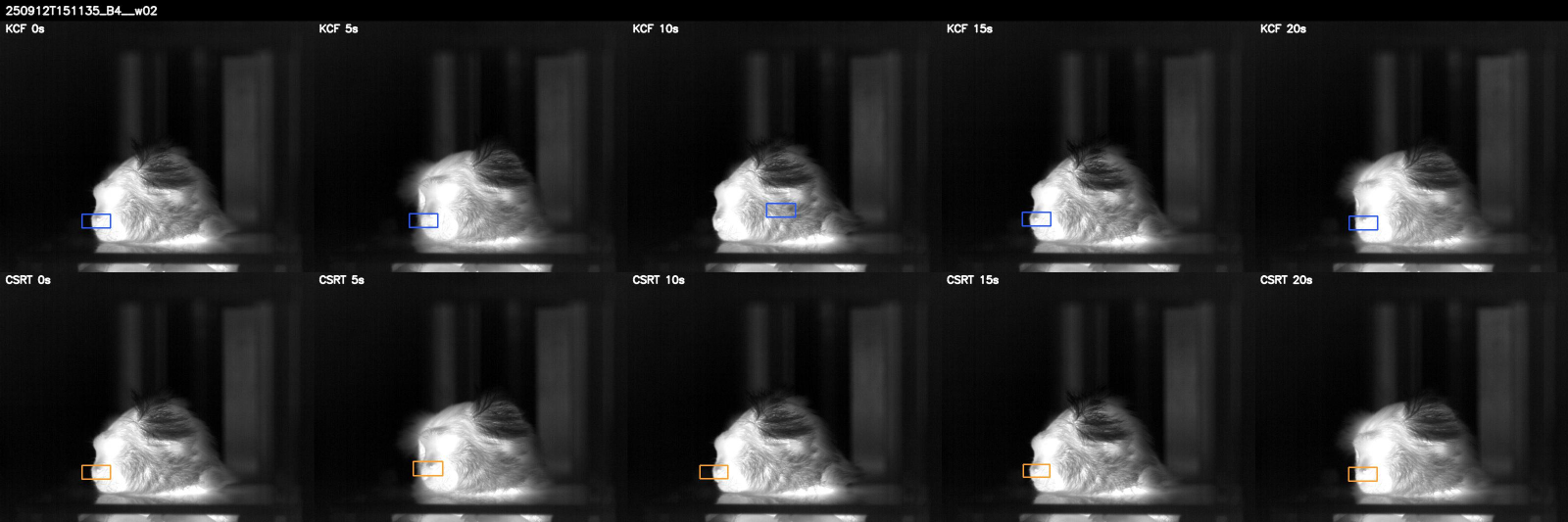}
        \caption{Example 2}
    \end{subfigure}
    \caption{Additional KCF and CSRT tracking comparisons under head-free
        motion.}
    \label{fig:appendix_kcf_csrt_tracking}
\end{figure*}

\begin{figure*}[!htbp]
    \centering
    \begin{subfigure}[t]{\linewidth}
        \centering
        \includegraphics[width=\linewidth]{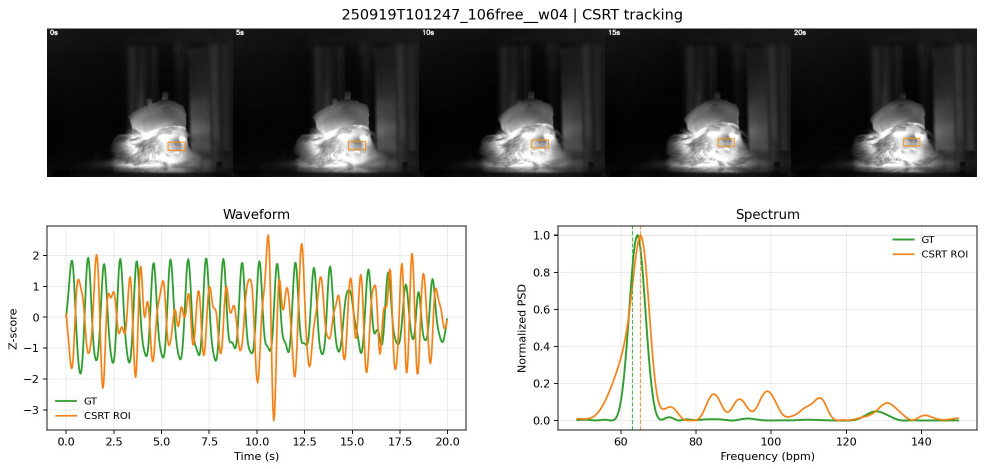}
        \caption{Successful case}
    \end{subfigure}
    \vspace{0.6em}

    \begin{subfigure}[t]{\linewidth}
        \centering
        \includegraphics[width=\linewidth]{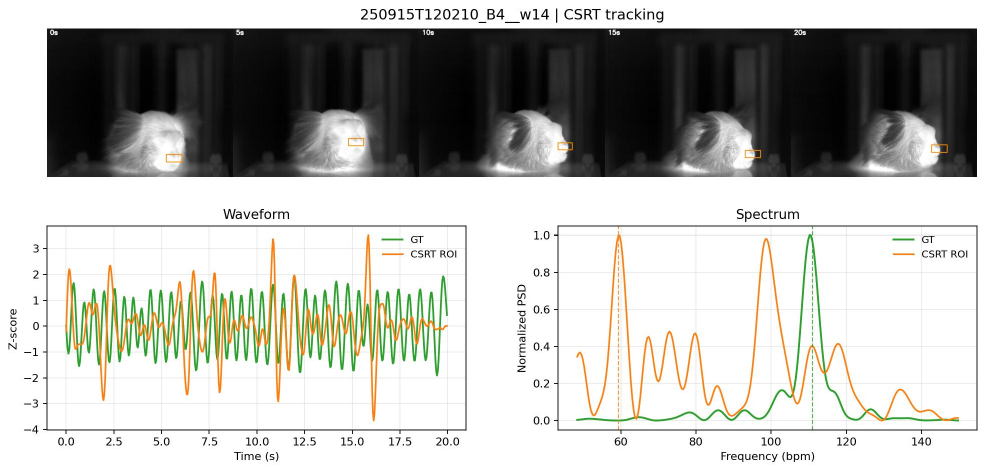}
        \caption{Failure case}
    \end{subfigure}
    \caption{Additional examples of tracked ROI respiratory rate estimation
        under head-free motion. Visual ROI continuity is not sufficient to
        guarantee an accurate respiratory trace.}
    \label{fig:appendix_tracking_cases}
\end{figure*}
\FloatBarrier

\end{document}